\documentclass[preprint,10pt,authoryear]{elsarticle}
\usepackage[margin=1in]{geometry}
\usepackage{hyperref}
\usepackage[utf8]{inputenc} 
\usepackage[T1]{fontenc}    
\usepackage{url}            
\usepackage{booktabs}       
\usepackage{amsfonts}       
\usepackage{nicefrac}       
\usepackage{microtype}      
\usepackage{lipsum}
\usepackage{amsmath}
\usepackage{amssymb}
\usepackage{amsthm}
\usepackage{bm}
\usepackage{url}
\usepackage{stackengine}
\usepackage{caption}
\usepackage{makecell}
\usepackage{multirow}
\usepackage{mwe}
\usepackage{subcaption}
\usepackage[most]{tcolorbox}
\usepackage{graphicx}
\usepackage{diagbox}
\usepackage{soul,color}
\usepackage[user,titleref]{zref}
\usepackage[ruled,vlined]{algorithm2e}
\usepackage{lineno}
\journal{}

\makeatletter
\def\ps@pprintTitle{%
  \let\@oddhead\@empty
  \let\@evenhead\@empty
  \let\@oddfoot\@empty
  \let\@evenfoot\@oddfoot
}
\makeatother

\begin{document}
\begin{frontmatter}







\title{Supply Chain Network Extraction and Entity Classification Leveraging Large~Language~Models}

\author[inst1]{Tong Liu}
\author[inst1]{Hadi Meidani}
\affiliation[inst1]{organization={University of Illinois, Urbana-Champaign, Department of Civil and Environmental Engineering},
            addressline={205 N Mathews Ave}, 
            city={Urbana},
            postcode={61801}, 
            state={IL},
            country={USA}}

\begin{abstract}
Supply chain networks are critical to the operational efficiency of industries, yet their increasing complexity presents significant challenges in mapping relationships and identifying the roles of various entities. Traditional methods for constructing supply chain networks rely heavily on structured datasets and manual data collection, limiting their scope and efficiency. In contrast, recent advancements in Natural Language Processing (NLP) and large language models (LLMs) offer new opportunities for discovering and analyzing supply chain networks using unstructured text data. This paper proposes a novel approach that leverages LLMs to extract and process raw textual information from publicly available sources to construct a comprehensive supply chain graph. We focus on the civil engineering sector as a case study, demonstrating how LLMs can uncover hidden relationships among companies, projects, and other entities. Additionally, we fine-tune an LLM to classify entities within the supply chain graph, providing detailed insights into their roles and relationships. The results show that domain-specific fine-tuning improves classification accuracy, highlighting the potential of LLMs for industry-specific supply chain analysis. Our contributions include the development of a supply chain graph for the civil engineering sector, as well as a fine-tuned LLM model that enhances entity classification and understanding of supply chain networks.
\end{abstract}


\end{frontmatter}


\section{Introduction}
\label{sec:introduction}

Supply chain networks are fundamental to the operational efficiency and competitiveness of industries worldwide \citep{pichler2023building,kosasih2024towards}. These networks comprise a complex web of interconnected entities, including suppliers, manufacturers, distributors, logistics providers, and customers. Each entity, represented as a node within the supply chain network, plays a distinct role, contributing to specific processes that collectively ensure the smooth functioning of the entire system \citep{habibi2023evaluating}. As industries continue to globalize, supply chains have expanded in both scale and complexity, presenting significant challenges in constructing these networks and identifying the roles of individual entities within them \citep{wu2023industry}. Effectively mapping these networks, a process known as supply chain network modeling, is crucial for minimizing costs and maintaining consistent product availability \citep{reyes2023development}. Furthermore, disruptions such as natural disasters and material shortages can cascade through supply chains, underscoring the importance of not only constructing robust networks but also understanding the intricate relationships and roles of the entities involved \citep{kanike2023factors,patel2023enhancing}. However, supply chain network modeling faces limitations related to information sharing and access. To illustrate, not all textural information is well-documented or readily available due to various factors, including unwillingness to share data, outdated documentation, and the multimodal nature of supply chain records. These challenges significantly hinder the accuracy and effectiveness of supply chain network modeling.

Multiple approaches have been proposed for modeling supply chain networks. Traditional rule-based approaches \citep{li2016rule, singh2019enhancing} model supply chain networks by establishing a set of rules designed by domain experts to address specific scenarios. While these approaches are relatively easy to implement and understand, they often rely heavily on expert knowledge and structured datasets, thus requiring significant human intervention for data management and analysis \citep{caiado2021fuzzy}. In contrast, ontology-based approaches represent knowledge through ontologies, which consist of concepts, relationships between concepts, and rules for reasoning \citep{kim2018toward}. These methods employ formal reasoning tools, such as semantic reasoning, to infer new knowledge \citep{ye2008ontology}. Compared to rule-based methods, ontology-based approaches offer greater adaptability to changes in the domain. However, inference and reasoning over complex domains and ontologies can be computationally expensive, and the formalization of ontologies may not always effectively capture domain-specific knowledge.

Advancements in Natural Language Processing (NLP) have created new opportunities for constructing and analyzing supply chain networks \citep{agarwal2019machine, wichmann2018towards}. NLP techniques such as word2Vec \citep{mikolov2013efficient}, TF-IDF \citep{aizawa2003information}, and transformers \citep{lin2022survey} enable capabilities like word embedding, information retrieval, and question answering, which facilitate the extraction of valuable insights from vast amounts of unstructured text data \citep{wichmann2020extracting}. Furthermore, these models can also be fine-tuned to extract specific relationships in different tasks, such as named entity recognition task \citep{naseer2021named, sharma2022named}. Nevertheless, NLP-based approaches face challenges in identifying hidden relationships between entities, particularly those not explicitly documented but critical within the network. Much of this information remains buried in multimodal sources. Addressing these challenges highlights a significant research gap in fully leveraging NLP for supply chain discovery and underscores the need for more advanced methods capable of handling unstructured data and revealing hidden connections between entities. In recent years, the internet has become a varied repository of supply chain data. Sources such as news articles, social media, and other publicly available platforms offer detailed insights into supplier-buyer relationships \citep{wichmann2020extracting}. However, this web-based information often exists in isolated, fragmented, and unstructured formats, making it difficult to identify and analyze the dependencies among these dispersed data points.

Recently, large language models (LLMs), trained on vast corpora of text, have demonstrated the ability to identify and map relationships between different entities by processing unstructured data and generating insightful connections \citep{chang2024survey}. LLMs offer a new direction for causal inference and question-answering \citep{yang2023understanding,li2024flexkbqa}. Unlike traditional NLP techniques, LLMs can perform zero-shot or few-shot learning, enabling them to generalize across various industries or contexts without requiring extensive retraining \citep{ahmed2022few}. Compared with other NLP techniques, a significant advantage of LLMs is their ability to uncover hidden or implicit relationships between entities that may not be explicitly stated in the text. For example, an LLM can analyze a broad array of text sources including news articles and social media posts to extract supply chain information such as company relationships and logistics dependencies with minimal manual intervention \citep{yang2024harnessing}. Furthermore, fine-tuning LLMs for domain-specific tasks allows for more precise and accurate predictions, providing new opportunities for building the supply chain networks \citep{ge2024openagi}.

In response to the increasing availability of internet-based textual information and the advancements in large language models, we propose a novel methodology that leverages LLMs to extract raw textural information from the internet to construct a supply chain graph for a specific industry. This paper uses the civil engineering industry as a case study to demonstrate the effectiveness of our approach. Furthermore, we present the development of a comprehensive supply chain graph for this sector, highlighting the efficacy of LLMs in extracting and structuring relevant information. Moreover, we fine-tune the LLMs to classify entities within the supply chain, providing deeper insights into the roles and interactions of various actors within the network. Through this process, we illustrate the potential of LLMs to enhance supply chain analysis by automating the discovery and categorization of entities across diverse information sources. Our main contributions are as follows: (1) This is the first study to leverage LLMs to automate the extraction of raw textual information from the Internet for supply chain network modeling and analysis; (2) We fine-tune the LLMs to classify and predict the categories of entities within the supply chain network, enhancing the granularity and depth of insights into supply chain dynamics; (3) This work establishes a groundwork for  future application of LLMs in industry-specific supply chain optimization and further exploration of NLP techniques for analyzing and improving supply chain networks.

The remainder of this article is structured as follows. Section \ref{sec:architecture} includes the explanation of the proposed pipeline for supply chain network modeling and entity classification. Furthermore, the case study with the civil engineering industry with the performance of entity classification of the proposed framework is discussed in Section \ref{sec:experiment}. Finally, the conclusion and discussion of the proposed framework are presented in Section \ref{sec:conslusion}.

\section{Proposed Methodology}
\label{sec:architecture}

The methodology for constructing a supply chain graph from raw textual data is divided into four key stages: data collection, prompt engineering, graph construction, and entity classification. Each stage addresses a specific component of the overall problem, integrating the capabilities of large language models for information extraction and graph representation. The approach is designed to build a scalable and accurate representation of inter-company relationships based on information extracted from online news sources.

\begin{figure}[htb!]
\centering
    \includegraphics[width=\textwidth]{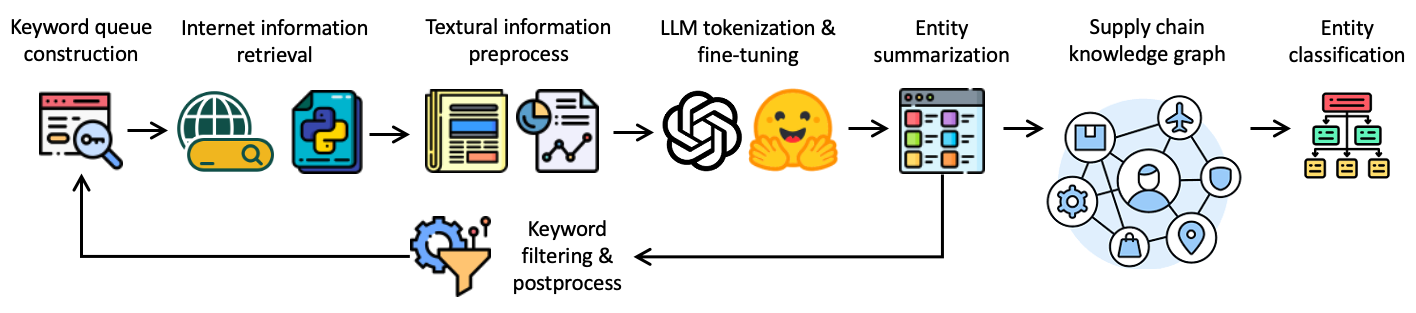}
    \caption{Workflow for supply chain graph construction using LLMs. The process begins with keyword queue construction, followed by information retrieval and text preprocessing. After fine-tuning, the LLM is used for entity classification and question answering, ultimately generating the supply chain network}
    \label{fig:model}
\end{figure}

\subsection{Data collection}

The data collection is done in an iterative way. We will create a queue of keywords we need to search. The graph construction phase begins with a queue that contains all identified entities, which will be considered as the node the supply chain graph, which will be explained in the following section.

The first stage of our methodology involves collecting relevant textual data from the news articles on the public domains. News articles are directly parsed from online sources, with the title and main content extracted and stored. To maintain a comprehensive view of each company's activity over time, we select at least ten news articles per year from 2018 to 2023 for each company. This results in a minimum of 50 raw textual data points per company. These text segments serve as the raw input for subsequent processing stages.

A pre-processing phase is then applied to normalize the textual data, which is critical for improving the accuracy of downstream tasks. All textual content, including keywords and company names, is converted to lowercase to ensure consistency and mitigate case-sensitivity issues. Additionally, a stopword list tailored for company names is maintained to remove common terms such as "Corp." or "Ltd." that do not contribute meaningful information. This stopword list is distinct from the one used for general textual information, and it is specifically designed to reduce noise in the data, thereby simplifying keyword matching in later stages.

Once preprocessing is complete, the textual information is incorporated into carefully designed prompts. Prompt engineering plays a crucial role in the information extraction process, as it leverages LLMs to derive meaningful insights from a raw text. In our approach, we employ two distinct types of prompts, each tailored to serve a specific function. Furthermore, we apply a chain-of-thought reasoning process to enhance the precision and depth of information extraction. Once the prompt is constructed, it is fed into the tokenizer. The tokenizer plays a crucial role by converting the input text into a sequence of tokens, which are the fundamental units that LLMs process. Tokenization ensures that the language model can interpret and handle the input efficiently. The outputs generated from these prompts are integral to downstream tasks, including network modeling and entity classification.

\subsection{Article Entity Summarization and Extraction}
\label{sec:prompt1}
In the initial phase of prompt engineering, we design prompts to ask the LLM for concise summaries of each company's nature and operations. One of the key motivations behind this prompt is the observation that if two companies are mentioned in the same article, it is highly likely they belong to the same industry, regardless of the type of relationship between these two companies. This process aids in extracting essential attributes, such as industry classification, primary products, and strategic roles within the supply chain. The prompt guides the LLM to generate summaries based on keywords and parsed content from relevant news articles. After extracting the relevant entities, we proceed to perform entity classification and relation extraction as downstream tasks.

For this task, we employ a few-shot learning approach, establishing a fixed format for the output within the prompt. The model is provided with input data and prompted to generate outputs according to the specified format. The prompt structure consists of three components: a system prompt, a user prompt, and a question prompt. Notably, each of these components can be adapted to accommodate a variety of supply chain-related tasks. The prompts are formulated as follows:

\begin{tcolorbox}[colback=black!5!white, colframe=black, fonttitle=\bfseries, title=Prompt for Article Entity Summarization, coltitle=white, colbacktitle=black]
\textbf{System Prompt} \\
You are a professional expert in the supply chain industry.\\
You only have one chance to answer the question. Always answer in the pattern:\\
1. [company name]: [short description] \\
2. [company name]: [short description] \\
3. [company name]: [short description] \\

\textbf{User Prompt} \\
Here is the news article about a company from the internet: \{news\} \\

\textbf{Question Prompt} \\
Give me the list and associated short description of the entity in \{industry\}
mentioned in the article.
\end{tcolorbox}

The variables within the curly brackets, "news" and "industry," represent dynamic entities that can be substituted with other industries or news sources. This adaptability allows the model to flexibly extract information and construct supply chain networks across different graphs.

For longer texts that exceed the token limit of the LLM, the input is segmented into smaller sections. This ensures that each segment remains within the model's token capacity, enabling effective processing without losing coherence or important information. These smaller sections are processed independently and then reassembled to maintain continuity and provide comprehensive insights for building supply chain graphs.

\subsection{Supply Chain Network Construction}
After extracting entities from the raw textual information, we proceed to construct the supply chain graph. For each keyword in the queue, and the corresponding entities identified by the language model, as described in Section \ref{sec:prompt1}, we establish links between the keywords and the relevant entities. Specifically, if two companies are mentioned together in an article, an edge is created between them, symbolizing a potential supply chain link or partnership. The entity extraction process is iterative, meaning the supply chain network evolves and expands progressively as new information is incorporated.

In the supply chain network, each node in the graph represents an entity in the industry. As new entities are identified during the entity extraction phase, they are added to the keyword queue for further processing and analysis. This iterative process continues until no additional company is found, ensuring a thorough exploration of the entire corpus of parsed documents.

To ensure the accuracy and consistency of the graph, a post-processing step is performed. Company names can appear in various forms, such as acronyms, abbreviations, or misspellings, which may lead to the creation of redundant nodes for the same entity. To address this, a continuously updated hashmap is maintained, mapping all detected variations of a company name to a single canonical form. If a duplicate is detected during the graph-building process, the redundant node is merged with the existing one, and the iteration proceeds. This mechanism not only prevents duplication but also enhances the robustness of the supply chain network representation.

\subsection{Supply Chain Entity Classification}
\label{sec:prompt2}
The second phase of prompt engineering extends the capabilities of  LLMs to identify the category of the detected entity, based on its summarization. In this phase, it is also crucial to determine both the nature of the entity and its associated business activities.

To construct the prompt, we follow a similar approach as outlined in Section \ref{sec:prompt1}, incorporating three key components into the prompt design. Given that large language models are causal in nature, we present the model with yes-or-no questions, guiding it to determine whether the entity belongs to a specific category. This structured questioning helps refine the categorization process and enhances the precision of the model's responses. The prompts are formulated as follows:

\begin{tcolorbox}[colback=black!5!white, colframe=black, fonttitle=\bfseries, title=Prompt for Entity Classification, coltitle=white, colbacktitle=black]
\textbf{System Prompt} \\
You are an experienced researcher proficient in entity labeling. \\
Your job is to determine whether the entity belongs to a particular category.
Answer with either 'Yes' or 'No', based on a description.\\

\textbf{User Prompt} \\
This is the description for \{company\_name\}: \{company\_description\} \\

\textbf{Question Prompt} \\
Is this entity \{company\_name\} mentioned in the description 
belonging to a \{company\_category\}? 
\end{tcolorbox}

In this task, the variables "company\_description," "company\_name," and "company\_category" are dynamic inputs that vary across different instances. Based on these inputs, we leverage LLMs to analyze the company description and subsequently classify the entity into a specific category. This classification process is crucial for accurately identifying the nature of the company. To further enhance the performance of the model, we implement a fine-tuning process. This fine-tuning is necessary because the model requires domain-specific knowledge to correctly classify entities within specialized categories. Fine-tuning allows the LLMs to incorporate contextual and industry-specific information, improving the accuracy of the classification, particularly in less common or complex cases. The details of the fine-tuning procedure, including the dataset selection, training process, and domain adaptation techniques, will be elaborated in the following section.

This entity classification step aims to categorize each company into one or more relevant classes, such as "supplier", "distributor", or "manufacturer". It is noted that a single entity can belong to multiple categories, we reformulate the multi-class classification problem into multiple binary classification tasks. For each class, a binary classifier is trained to determine the presence or absence of that category for a given company.

\section{Numerical Experiments}
\label{sec:experiment}

\subsection{Experiment Setup}
\label{sec:setup}
To assess the effectiveness of our proposed method for extracting and structuring supply chain information using LLMs, we conducted a comprehensive case study within the civil engineering industry. This sector presents a complex and dynamic landscape with multiple interconnected domains \cite{liu2023physics,liu2024graph}, making it an ideal test bed for evaluating the robustness, adaptability, and scalability of our approach. Civil engineering encompasses various critical areas, such as construction, infrastructure development, transportation systems, and materials. These domains generate significant volumes of unstructured data that are essential for supply chain analysis but present substantial challenges for extraction and management using traditional techniques.

Our study commenced with the collection of data from publicly available news sources. Specifically, we retrieved the news articles from the Google news platform. To ensure the dataset’s comprehensiveness and relevance, we retrieved articles from major international news sources, covering a broad temporal scope. The resulting dataset comprises news articles spanning infrastructure projects, supply chain disruptions, procurement strategies, and regulatory shifts pertinent to civil engineering firms. To streamline our analysis, we categorized the civil engineering industry into distinct sectors, as illustrated in Table \ref{tab:category}, allowing for a more targeted and systematic approach to information extraction. This categorization not only enhanced the granularity of our analysis but also enabled the application of domain-specific extraction rules tailored to the unique challenges and characteristics of each sector.

Subsequent to data collection and categorization, we employed multiple state-of-the-art LLMs to extract and structure the information contained within the news articles. These models were selected for their advanced capabilities in handling diverse and unstructured textual data. The LLMs were fine-tuned to detect and extract relevant supply chain entities, which were then systematically mapped to construct supply chain graphs. To extract and structure the information from the collected articles, we employed multiple state-of-the-art large language models. Specifically, we utilized the following models:

\begin{itemize} 
    \item \textbf{LLaMA}: A powerful transformer-based language model, optimized for multilingual tasks, particularly effective in analyzing and disambiguating complex linguistic structures found in news reports and technical documents. 
    \item \textbf{LLaMA2}: An enhanced version of LLaMA, offering improved accuracy and computational efficiency. LLaMA2 was leveraged for more precise extraction of entities and relationships, including context-aware disambiguation of overlapping or ambiguous information.
    \item \textbf{Mistral}: A highly specialized model, fine-tuned for domain-specific text, particularly adept at processing technical and industry-related content. Mistral played a critical role in extracting deeper, context-specific insights and identifying nuanced relationships within specialized datasets.
\end{itemize}

\begin{table}[htb!]
\caption{List of entity categories for supply chain network, each category is associated with an example of a company and the summarization}
\label{tab:category}
\begin{tabular}{m{0.15\linewidth}m{0.2\linewidth}m{0.55\linewidth}}
\hline
Category & Example & \multicolumn{1}{l}{Summarization} \\ \hline
\begin{tabular}[l]{@{}l@{}}Engineering\\consulting \end{tabular} & Parsons Corporation & A global engineering, construction, technical, and management services   company with a focus on infrastructure, defense, and intelligence \\
\begin{tabular}[l]{@{}l@{}}Construction\\contractor \end{tabular}  & Yates Construction & A construction company based in Philadelphia, Mississippi, part of the team that will lead the construction work for the honda ev battery factory project in ohio \\
\begin{tabular}[l]{@{}l@{}}Material\\supplier \end{tabular}  & National Lime Stone & A global leader in limestone solutions for industry and agriculture \\
\begin{tabular}[l]{@{}l@{}}Government\\agency \end{tabular}  & TxDOT & An abbreviation for the texas dept. of transportation, the state agency   responsible for planning, constructing, and maintaining the state’s   transportation infrastructure \\
\begin{tabular}[l]{@{}l@{}}Equipment\\lessor \end{tabular} & Bragg Crane Service & A crane service subcontractor \\
\begin{tabular}[l]{@{}l@{}}Insurance\\provider \end{tabular} & Amtrust Financial & A global provider of specialty insurance products and services \\
\begin{tabular}[l]{@{}l@{}}Real estate\\developer \end{tabular} & Lendlease & An Australian real estate investment giant that provides services in   property development, construction, and investment management \\
\begin{tabular}[l]{@{}l@{}}Legal\\counsel \end{tabular} & McGuireWoods & A full-service law firm with a strong focus on government contracts and   national security \\
\begin{tabular}[l]{@{}l@{}}Software\\service \end{tabular} & Aconex & A software company that provides project management and collaboration   tools for the construction industry \\ \hline
\end{tabular}
\end{table}

To tailor pre-trained Large Language Models (LLMs) to our specific task in the civil engineering domain, we implemented a series of Parameter-Efficient Fine-Tuning (PEFT) strategies. These methods allowed us to effectively adapt the models without the need for extensive retraining, optimizing computational resources while maintaining high levels of performance. The fine-tuning process involved the following steps:

\begin{itemize} 

\item \textbf{Domain-Specific Corpus Augmentation:} We enhanced the pre-trained LLMs by augmenting their training data with a carefully curated set of civil engineering-specific technical documents, white papers, and industry reports. This augmentation enabled the models to better capture domain-specific terminology and contextual nuances within the civil engineering sector.

\item \textbf{Adapter-Based Fine-Tuning:} To efficiently fine-tune the models without significantly increasing computational costs, we incorporated adapter modules. These adapters learned task-specific parameters while keeping most of the original pre-trained model’s weights frozen, ensuring that the fine-tuning process was both cost-effective and robust across tasks.

\item \textbf{Low-Rank Adaptation (LoRA):} As part of our PEFT strategy, we implemented LoRA, a method that factorizes the weight update matrices within the transformer layers. This approach significantly reduced the number of trainable parameters, allowing us to achieve effective model adaptation with minimal computational overhead.
\end{itemize}

Following the fine-tuning process, the adapted LLMs were deployed to extract key entities and relationships from unstructured text sources, such as news articles, technical reports, and industry updates. We focused on identifying entities such as company names, project locations, material suppliers, and logistical partners. In parallel, we extracted relationships reflecting supply chain dependencies, project timelines, procurement agreements, and collaborative ventures. This information was systematically used to construct a supply chain graph tailored to the civil engineering industry. In this graph, nodes represented entities (e.g., companies, projects, or suppliers), while edges captured the interactions or relationships between them (e.g., supply contracts, project collaborations, or partnership agreements). Finally, we validated the graph's accuracy through domain expert review and iterative refinement, ensuring its relevance and utility in understanding the dynamics of supply chains in civil engineering projects.

\subsection{Experiment Result}

The foundational component of our research is the construction of a graph representing the supply chain relationships within the civil engineering sectors. This graph was built from raw text data extracted from the internet. Using the proposed pipeline as discussed in Section \ref{sec:architecture}, we identified key entities such as companies, projects, materials, and government agencies, and established relationships among them.

\subsubsection{Supply Chain Network Construction}

\begin{figure}[htbp]
\centering
\begin{minipage}{0.45\textwidth}
\centering
\includegraphics[width=\linewidth]{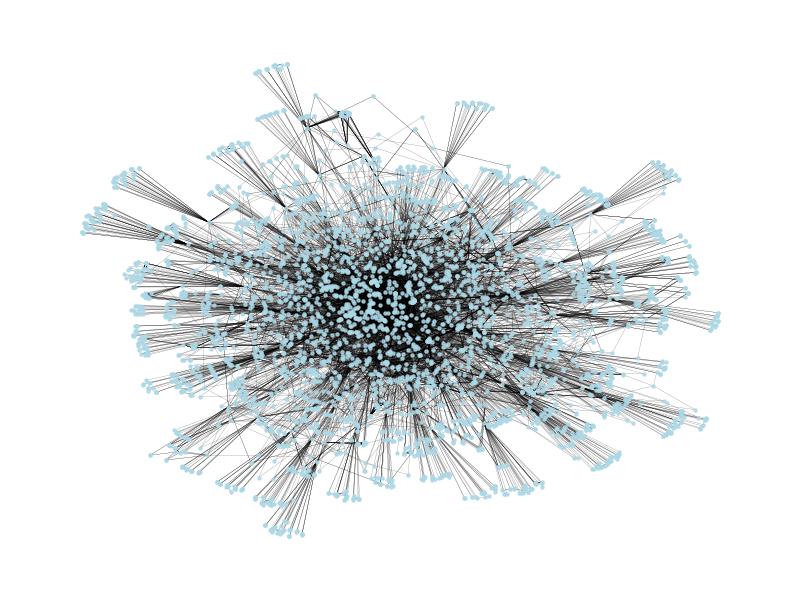} 
\caption{Constructed supply chain network graph. Each node in the graph represents a distinct entity in the civil engineering industry.}
\label{fig:graph}
\end{minipage}
\hfill
\begin{minipage}{0.45\textwidth}
\centering
\includegraphics[width=\linewidth]{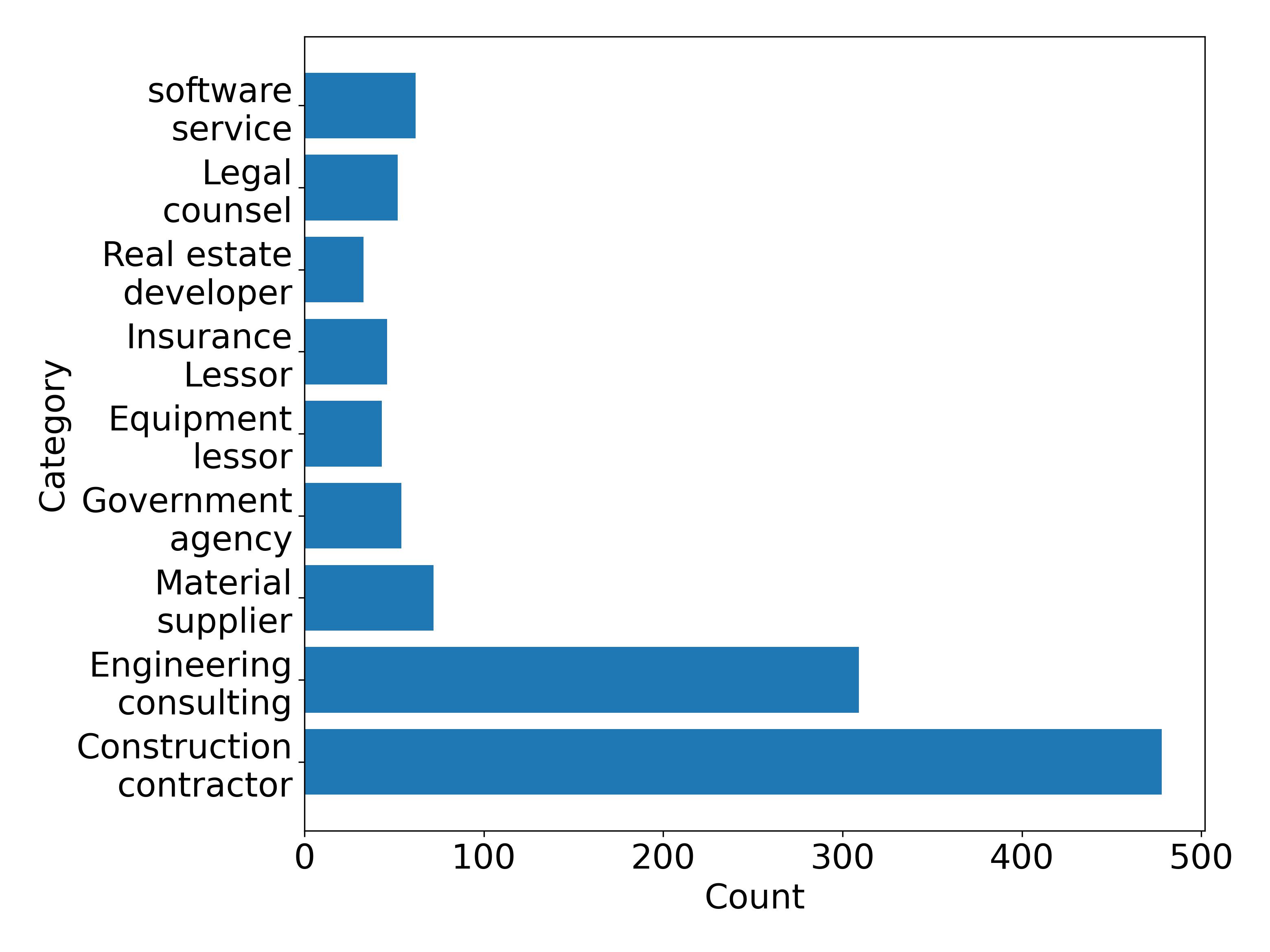} 
\caption{Distribution of the categories in the supply chain network graph. The labelled data will be used in the fine-tuning LLMs}
\label{fig:bar}
\end{minipage}
\end{figure}

The visualization of the overall supply chain network is presented in Figure \ref{fig:graph}. The resulting graph contains 4,293 nodes and 16,793 edges. Nodes represent key entities, including companies, government bodies, and materials, while edges capture the relationships and interactions between these entities. The graph spans multiple sectors, encapsulating various components of the civil engineering and infrastructure supply chain. Notably, each node is highly interconnected with other nodes, indicating the complex interdependencies that exist within the network. For instance, in the civil engineering industry, no single company can typically complete a large-scale project independently. Rather, the successful execution of such projects relies on the collaboration of multiple entities from different subdomains, each contributing specialized expertise and resources.

\begin{figure}[htb!]
\centering
\begin{subfigure}[b]{0.49\textwidth}
    \centering
    \includegraphics[width=\textwidth]{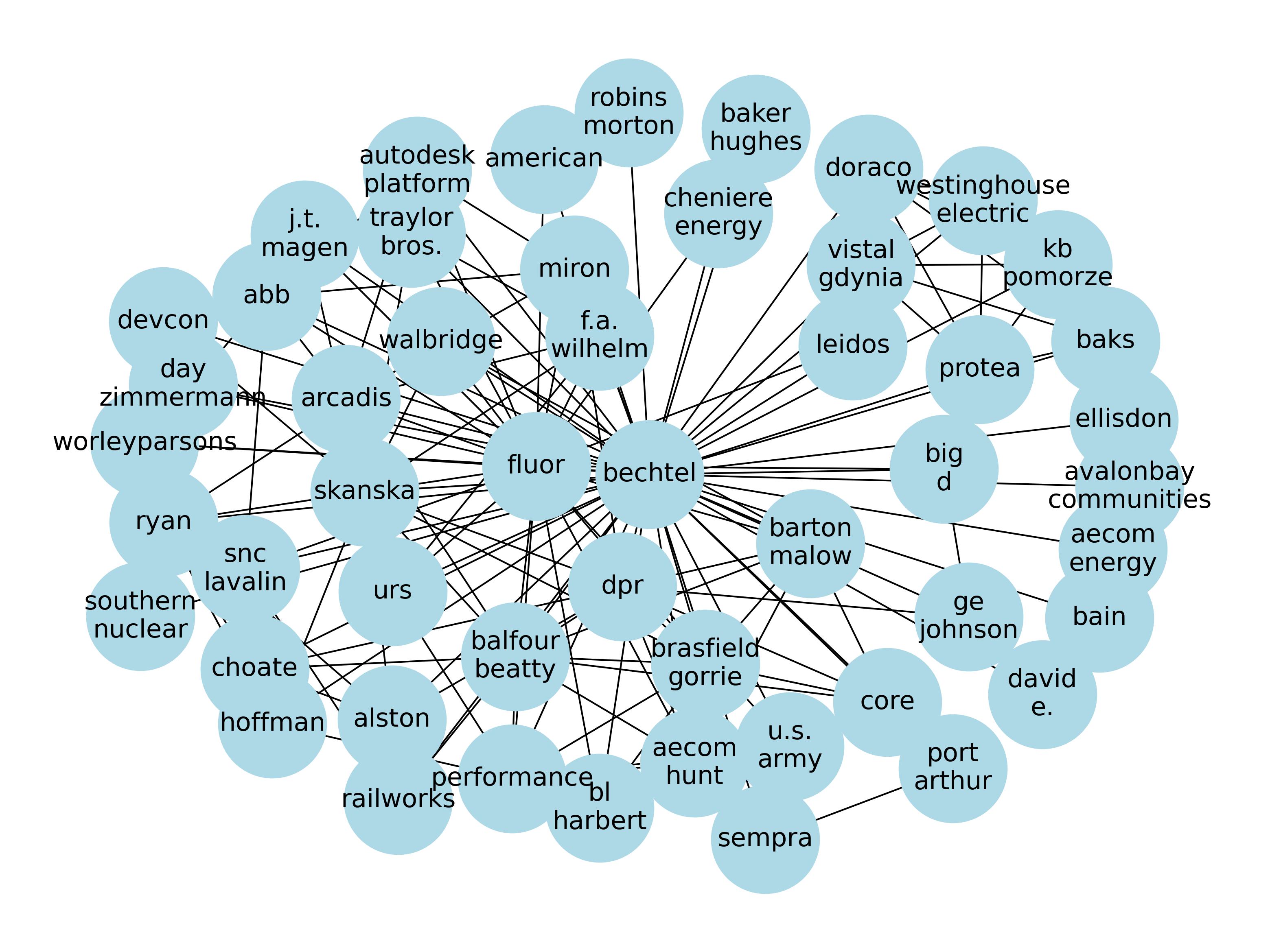}
    \caption{Bechtel}
    \label{fig:subgraph_bechtel}
\end{subfigure}
\hfill
\begin{subfigure}[b]{0.49\textwidth}
    \centering
    \includegraphics[width=\textwidth]{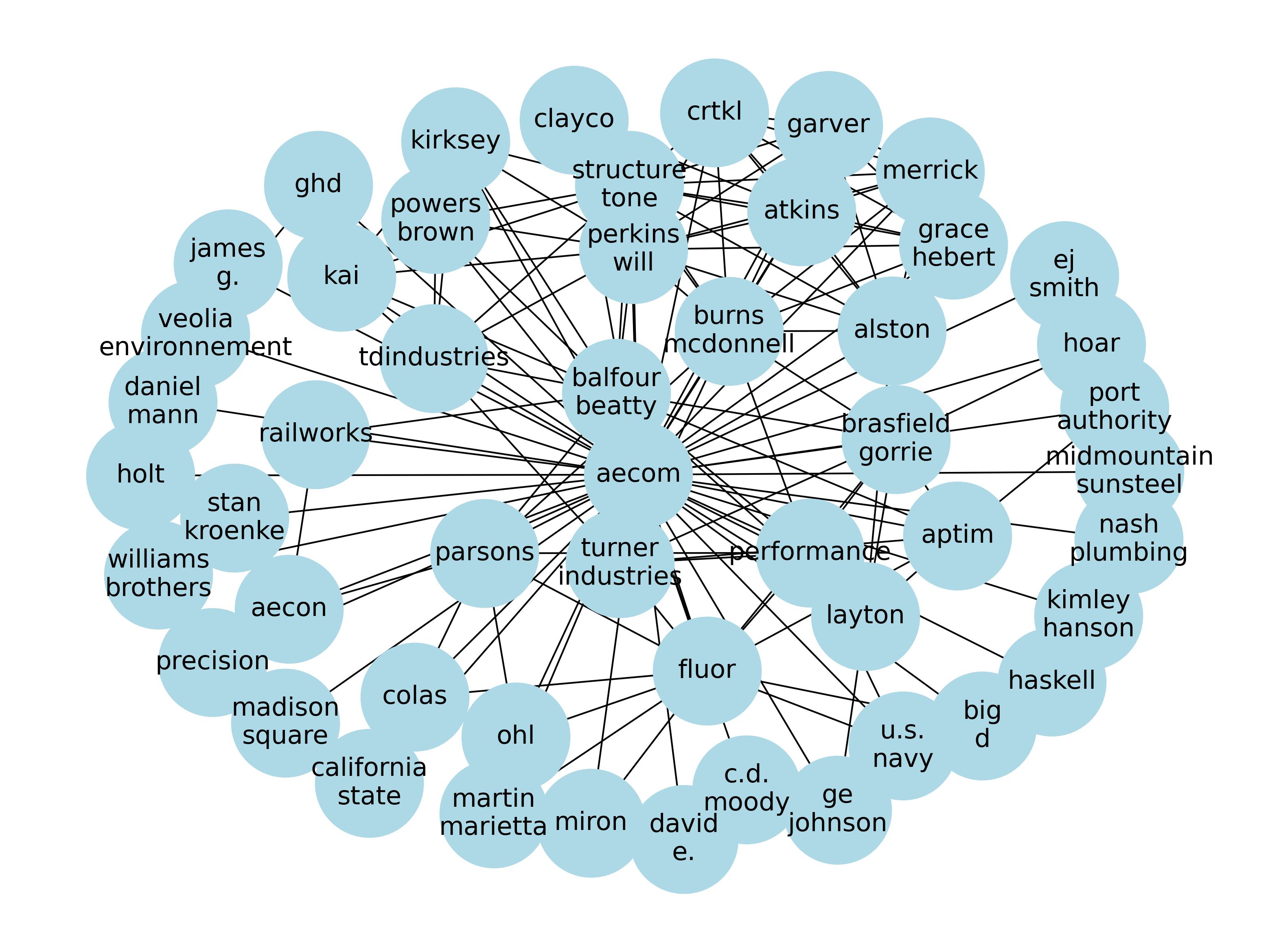}
    \caption{AECOM}
    \label{fig:subgraph_aecom}
\end{subfigure}
\caption{Subgraph constructed from the single keyword. We take AECOM and Bechtel as two examples, for each grpah, we sampled 50 neighbors and related entity to visualize.}
\label{fig:subgraph}
\end{figure}

This interdependence is further illustrated through more focused visualizations, as shown in Figures \ref{fig:subgraph}. These subgraphs \ref{fig:subgraph_bechtel}, and \ref{fig:subgraph_aecom} highlight the specific supply chain networks of companies like Bechtel, focusing on the 1-hop neighbors—entities that are directly connected to the focal company. Although the original subgraph of Bechtel includes over 200 nodes, for clarity, we sampled 50 entities directly related to the keyword "Bechtel." A similar approach is taken with AECOM’s subgraph. From these subgraphs, it becomes evident that these companies are embedded in highly interconnected networks where their operations depend on various partners and collaborators. The high density of connections underscores the critical role of inter-organizational collaboration in achieving project success across the supply chain.

\subsubsection{Entity Classification}
After constructing the supply chain network and completing the entity extraction and summarization tasks, the next phase involves classifying the entities within the network. As detailed in Section \ref{sec:setup}, we leveraged several state-of-the-art large language models (LLMs) to perform the classification. Specifically, we employed LLaMA-7B, LLaMA2-7B, and LLaMA2-13B as the backbone models for this task.

Although these LLMs are pre-trained on extensive corpora, they were further fine-tuned on domain-specific knowledge relevant to the supply chain context. To improve the performance of the models on this task, we fine-tuned them using the annotated dataset, incorporating domain-specific prompts as outlined in Section \ref{sec:prompt2}.  For the entity classification task, we curated and manually annotated a dataset, ensuring that each entity was assigned to the correct category based on its role within the network. The distribution of labeled entities across various categories is presented in Figure \ref{fig:bar}, where it is evident that the majority of the data belongs to the categories of `` Engineering Consulting" and ``Construction Contractor". The fine-tuning process involved adjusting the model's parameters based on the labeled data, optimizing it to better understand the nuances of the supply chain categories. The details of the fine-tuning procedure are discussed in the following section.

To evaluate the performance of the entity classification model, we conduct several experiments under different scenarios. These scenarios are designed to assess the model's ability to handle both raw and processed data, as well as to investigate the impact of model fine-tuning on classification performance. The specific scenarios are as follows: (1) applying the model directly to the question-answering task using the raw text articles without any preprocessing; (2) applying the model to the same task using a summarized version of the text, thus reducing the input size; (3) fine-tuning the model on the entire labeled dataset to improve its generalizability; (4) fine-tuning the model on a balanced subset of the data, obtained through downsampling, to mitigate any potential bias due to class imbalance.

For model performance evaluation, we use two key metrics: classification accuracy and the F1 score. Accuracy provides an overall measure of correctness, while the F1 score accounts for the balance between precision and recall, particularly important in cases of class imbalance. The mathematical formulation of these metrics is presented as follows:

\begin{equation}
\label{ref:accuracy}
\text{Accuracy} = \frac{TP + TN}{TP + TN + FP + FN},
\end{equation}

\begin{equation}
\label{ref:F1}
F_1 = 2 \cdot \frac{\text{Precision} \cdot \text{Recall}}{\text{Precision} + \text{Recall}} = \frac{2TP}{2TP + FP + FN}.
\end{equation}

The results, detailed in Table \ref{tab:performance}, demonstrate the effectiveness of fine-tuning large language models (LLMs) for classifying entities within a supply chain graph. Specifically, the fine-tuned model significantly outperforms its original counterpart, achieving higher F1 scores across all classes, which indicates a marked improvement in capturing domain-specific knowledge and relationships. This finding highlights the benefit of fine-tuning the model on industry-relevant data, enabling it to generalize better to tasks related to civil engineering and infrastructure.

We evaluated three variants of the LLaMA model (LLaMA-7b, LLaMA2-7b, and LLaMA2-13b) under different training and evaluation conditions: raw text input, text summarization, and fine-tuning on both unbalanced and balanced datasets. The results reveal the progression of performance across these models and tasks.

\begin{table}[htb!]
\centering
\caption{Model performance comparison of entity classification for supply chain network graph with different question-answering setting.}
\label{tab:performance}
\resizebox{\textwidth}{!}{%
\begin{tabular}{cccccc}
\hline
Model & Metric & w/ raw article & w/ summarizaton & Finetune   w/ unbalanced data & Finetune   w/ balanced data \\ \hline
\multirow{2}{*}{LLaMA-7b} & Accuracy & 0.591 & 0.621 & 0.889 & 0.929 \\
 & F1 score & 0.283 & 0.315 & 0.601 & 0.693 \\
\multirow{2}{*}{LLaMA2-7b} & Accuracy & 0.662 & 0.662 & \textbf{0.951} & 0.948 \\
 & F1 score & 0.323 & 0.323 & \textbf{0.752} & 0.714 \\
\multirow{2}{*}{LLaMA2-13b} & Accuracy & \textbf{0.895} & \textbf{0.895} & 0.951 & \textbf{0.958} \\
 & F1 score & \textbf{0.599} & \textbf{0.599} & 0.631 & \textbf{0.765} \\ \hline
\end{tabular}%
}
\end{table}

For the raw text input scenario, the LLaMA-7b model achieved an accuracy of 0.591 and an F1 score of 0.283, demonstrating moderate accuracy but low balance between precision and recall. The LLaMA2-7b model slightly improved on these results, with an accuracy of 0.662 and an F1 score of 0.323. However, the LLaMA2-13b model exhibited the best performance, with an accuracy of 0.895 and an F1 score of 0.599, suggesting that larger models are more capable of handling raw input data.

When summarization was applied to the input, accuracy for LLaMA-7b and LLaMA2-7b remained similar to the raw text results, but their F1 scores showed slight improvement. This suggests that summarization aids in balancing precision and recall without significantly affecting overall accuracy. The LLaMA2-13b model maintained its strong performance with no notable gains from summarization, likely due to its capacity to process raw data effectively.

Fine-tuning on the unbalanced dataset led to significant improvements across all metrics. The LLaMA-7b model’s accuracy rose to 0.889, with an F1 score of 0.601, demonstrating its enhanced ability to classify minority classes. The LLaMA2-7b model reached an accuracy of 0.951 and an F1 score of 0.752, showing strong gains. The LLaMA2-13b model also achieved 0.951 accuracy, though with a slightly lower F1 score of 0.631, which might indicate some challenges with class imbalance. Further improvements were observed when fine-tuning was performed on a balanced dataset, achieved via downsampling. The most substantial gains were seen in the F1 scores, indicating improved classification performance across all classes. The LLaMA2-13b, the best-performing model, achieved an accuracy of 0.958 and an F1 score of 0.765, demonstrating the effectiveness of balanced fine-tuning, particularly in terms of achieving better precision and recall. In the fine-tuning process, we fine-tuned 10\% of the model's parameters, specifically focusing on layers responsible for capturing relationships between entities. This underscores the importance of domain-specific fine-tuning in improving model performance on specialized tasks.

\section{Conclusion and Discussion}
\label{sec:conslusion}

In this paper, we have demonstrated the utility of constructing a domain-specific supply chain graph by utilizing large language models for entity extraction and classification. The resulting graph, encompassing tens of thousands of entities and relationships, provides valuable insights into the civil engineering industry’s complex network of companies, projects, and government bodies. By employing both a general pre-trained LLM and a fine-tuned version of the model, we found that the fine-tuned model significantly outperforms the baseline in classifying entities, as indicated by higher F1 scores.

The improvement in classification accuracy emphasizes the importance of domain-specific adaptation when applying LLMs to real-world problems, particularly in highly specialized industries. This approach not only improves entity classification but also offers enhanced capabilities for understanding and optimizing supply chain networks. Future work could explore extending this methodology to other industries or incorporating additional types of relationships and entity attributes to further enrich the graph’s analytical power.

\bibliographystyle{elsarticle-harv} 
\bibliography{cas-refs}

\end{document}